\documentclass{article}

\usepackage{PRIMEarxiv}

\usepackage[utf8]{inputenc} 
\usepackage[T1]{fontenc}    
\usepackage{hyperref}       
\usepackage{url}            
\usepackage{booktabs}       
\usepackage{amsfonts}       
\usepackage{nicefrac}       
\usepackage{microtype}      
\usepackage{lipsum}
\usepackage{fancyhdr}       
\usepackage{graphicx}       
\usepackage{amsmath}        
\usepackage{svg}
\graphicspath{{media/}}     

\title{SentinelAI: A Multi-Agent Framework for Structuring and Linking NG9-1-1 Emergency Incident Data
\thanks{\textit{\underline{Citation}}: 
\textbf{Ho, K., Zaslavsky, I. SentinelAI: A Multi-Agent Framework for Structuring and Linking NG9-1-1 Emergency Incident Data.}} 
}

\author{
  Kliment Ho, Ilya Zaslavsky \\
  San Diego Supercomputer Center \\
  UC San Diego \\
  La Jolla, CA \\
  \texttt{\{kho005, izaslavsky\}@ucsd.edu} \\
}

\begin{document}
\maketitle

\begin{abstract}
Emergency response systems generate data from many agencies and systems. In practice, correlating and updating this information across sources in a way that aligns with Next Generation 9-1-1 (NG9-1-1) data standards remains challenging. Ideally, this data should be treated as a continuous stream of operational updates, where new facts are integrated immediately to provide a timely and unified view of an evolving incident. This paper presents SentinelAI, a data integration and standardization framework for transforming emergency communications into standardized, machine-readable datasets that support integration, composite incident construction, and cross-source reasoning. SentinelAI implements a scalable processing pipeline composed of specialized agents. The EIDO Agent ingests raw communications and produces NENA-compliant Emergency Incident Data Object JSON (EIDO-JSON). The Incident Data Exchange (IDX) Agent performs incident correlation across EIDOs, and the Geocoding Agent provides spatial enrichment. The contributions of this work are threefold: (1) the implementation of an agent-based reference architecture aligned with the recently adopted NENA EIDO standard; (2) the operational treatment of EIDO as an evolving incident representation that is updated incrementally as new information arrives; and (3) the demonstration of enterprise interoperability via integration with FME. This work focuses on data structuring and interoperability to support downstream systems.
\end{abstract}


\section{Introduction}
Emergency response involves many kinds of data, both structured and unstructured, produced by different agencies and systems \cite{fcc_ng911}. Information from operational systems, including dispatch platforms, reporting tools, communication systems, as well as news and governmental sources, often combines structured fields with unstructured elements such as free-form text and audio. This data remains difficult to integrate and process consistently at scale. Information needed to understand complex events (generated by police, fire, medical services, local governments, and news organizations) typically follows its own content and formatting standards and lifecycles. To enable cross-source analysis and interpretation of linked events, this information must first be transformed into a consistent, machine-readable, and standardized representation.

\subsection{The Next Generation 9-1-1 Context}
The deployment of Next Generation 9-1-1 (NG9-1-1) systems across the United States is changing emergency communications infrastructure. Unlike earlier systems that were primarily designed around voice calls and limited media types, NG9-1-1 is an IP-based system designed to handle multiple modes of communication, including text, images, video, and sensor data, from diverse devices and locations \cite{nena_i3, cisa_ng911}. This transition, now underway nationwide, increases both the diversity and volume of data flowing through Public Safety Answering Points (PSAPs).

To support interoperability between systems operated by different agencies and jurisdictions, the National Emergency Number Association (NENA) defined a standard architecture that includes functional elements and interfaces for next-generation emergency services \cite{nena_i3}. A recently published NENA-STA-024.1.1-2025 standard formally specifies the Emergency Incident Data Object (EIDO) format \cite{nena_eido}. EIDO provides a canonical JSON schema for representing emergency incidents, designed to facilitate seamless information exchange across the NG9-1-1 ecosystem.

Complementing EIDO, the Incident Data Exchange (IDX) concept addresses the challenge of correlating and aggregating multiple EIDOs that describe the same real-world emergency event. In modern multi-agency responses, a single incident may generate many data points from different sources, including initial 9-1-1 calls, dispatch updates, officer reports, fire service notifications, and emergency medical services records \cite{palen2016crisis}. IDX provides the conceptual framework for linking these discrete data objects into a composite, evolving representation of the incident.

In practice, incident information is distributed across multiple platforms, arrives incrementally, and varies in structure, terminology, and timing. Approaches that rely on monolithic data transformation pipelines often struggle to combine language interpretation, temporal correlation, spatial interpretation, and semantic normalization within a single workflow.

\subsection{Contributions}
This paper presents SentinelAI, a system that addresses these challenges by decomposing the complex objective of data structuring into discrete sub-tasks, each delegated to a specialized autonomous agent. This multi-agent approach decomposes the overall task into smaller, well-defined processing steps. Our primary contributions are:

\begin{enumerate}
    \item \textbf{Three-Agent Reference Architecture}: A concrete reference implementation composed of three specialized agents (EIDO Agent, IDX Agent, and Geocoding Agent) that coordinate asynchronously to structure, correlate, and spatially enrich incident information in alignment with the NENA EIDO standard.
    \item \textbf{Operational Linked-Incident Data Model}: A methodology for representing incidents where individual EIDO objects are linked to a common incident context, allowing incremental updates from multiple sources without collapsing information into a single record.
    \item \textbf{Enterprise Integration}: A demonstration of interoperability with existing geospatial and emergency management systems through integration with FME.
\end{enumerate}

The goal of SentinelAI is to support the consistent structuring and linkage of incident information so that it can be interpreted by both human operators and computational systems for analysis and operational use.

\section{The Emergency Incident Data Object Standard}
The EIDO-JSON format is the canonical data structure at the heart of the SentinelAI framework. The schema implements the NENA-STA-024.1.1-2025 standard for Next Generation 9-1-1 (NG9-1-1) data exchange \cite{nena_eido, nena_i3}. This section describes the core components of EIDO and places the standard in the context of existing event and observation data models.

\subsection{Conceptual Framework}
A key principle of the EIDO standard is the concept of a composite incident. Emergency events evolve as information arrives from multiple sources, such as initial calls, dispatch updates, unit status changes, and field reports. Each new piece of information is structured as a distinct EIDO-JSON document that captures the incident state at a given time. The IDX Agent links these documents to a common incident context, creating a linked representation of the incident rather than a single static record.

\begin{figure}[ht]
  \centering
  \includegraphics[width=1\linewidth]{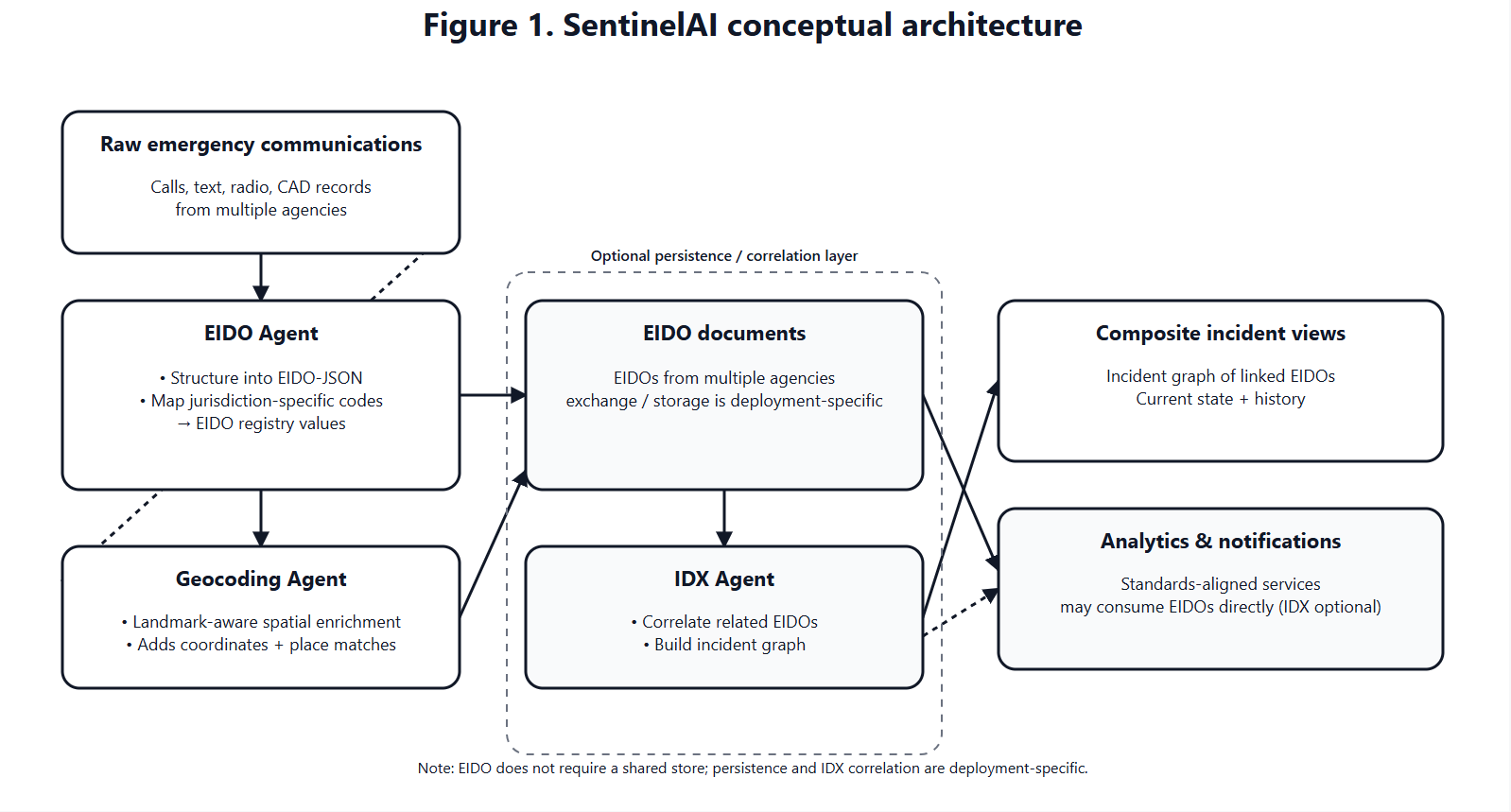}
  \caption{SentinelAI reference architecture showing the three-agent workflow. The EIDO Agent structures incoming reports into EIDO-JSON, the IDX Agent links related EIDOs into composite incidents, and the Geocoding Agent enriches location information. Agents interact through well-defined interfaces and support analytical and integration services.}
  \label{fig:arch}
\end{figure}

The overall design is shown in Figure \ref{fig:arch}. Incoming incident reports from multiple sources are first processed by the EIDO Agent, which produces standardized EIDO-JSON objects. These objects are persisted and made available to other system components. The IDX Agent operates over this store to associate related EIDOs with a common incident context and to maintain composite incident views. The Geocoding Agent enriches location information using external reference data and services. Together, the agents operate asynchronously, with each focusing on a well-defined responsibility while contributing to a unified incident representation.

This approach preserves how incident information accumulates over time while maintaining explicit links between related updates. Other systems can reconstruct incident history by following references between linked EIDO objects or retrieve the most recent state from the latest linked update. The design aligns with the operational realities of emergency response without imposing a rigid linear or graph-based incident model.

\subsection{EIDO-JSON Structure and Interoperability Principles}
The EIDO-JSON schema organizes incident data into discrete components, including core event details (Incident), responding units (Resource), associated persons and vehicles (Entity), communication events (such as Call), and multi-format spatial information (Location).

A core principle of the EIDO standard is its reliance on controlled vocabularies and registries managed by the Internet Assigned Numbers Authority (IANA) for key classification fields. This design ensures that incident types, priorities, and roles are interpreted consistently across agencies and jurisdictions, avoiding ambiguity introduced by local coding schemes. For example, different agencies may use conflicting codes such as ``Code 3'' or ``P1'' to denote high urgency; EIDO mandates mapping these to a standardized numeric priority scale to remove such ambiguities.

Unlike many geospatial and event standards that emphasize precise geometric representations, EIDO accommodates multiple forms of location description, including civic addresses, landmarks, free-text descriptions, and geographic coordinates. This flexibility reflects emergency communications practice, where locations are often reported using informal or contextual references that require interpretation rather than direct coordinate parsing. The standard provides sufficient structure for machine processing while preserving the way information is communicated during emergency response.

\section{SentinelAI Agents}
SentinelAI is implemented as a system composed of three specialized agents, each responsible for a distinct function in structuring, correlating, and enriching emergency incident data. Rather than relying on a single monolithic pipeline, the system decomposes processing into agent-level responsibilities, allowing components to operate independently and be composed flexibly. The following sections describe the behavior and responsibilities of each agent in detail.

\subsection{EIDO Agent: Structuring and Interpretation}
The EIDO Agent serves as the primary data ingestion point and is responsible for transforming heterogeneous input formats into standardized EIDO-JSON documents. The agent uses a language model to interpret and combine information from emergency communications that vary in format, wording, and level of detail. The specific model choice is not material to the architecture and is treated as a replaceable component.

The agent performs several critical interpretation tasks to normalize incoming data. First, it extracts structured information from unstructured sources such as call-taker notes, dispatch summaries, and radio transcripts, identifying entities like people, locations, and organizations. Second, it handles the translation of agency-specific codes; for example, translating local incident types like ``211A'' or priority codes like ``Code 3'' into the standardized IANA registry values required by EIDO. Third, the agent synthesizes timestamp information, combining legacy fields (which may split dates and times across multiple columns) into standard ISO 8601 timestamps so that events can be ordered reliably.

To ensure output consistency, the agent applies configurable EIDO templates that define which components and fields are expected for different categories of incidents. These templates distinguish required from optional information and are maintained outside application code, allowing subject-matter experts to adjust data requirements for specific incident types without modifying core system logic. Table \ref{tab:transformations} provides a worked example of how the EIDO Agent maps specific legacy CAD fields into the target EIDO-JSON structure.

\begin{table}[ht]
 \caption{Representative transformations performed by the EIDO Agent}
  \centering
  \begin{tabular}{p{0.25\linewidth} p{0.35\linewidth} p{0.3\linewidth}}
    \toprule
    Legacy CAD Field & EIDO-JSON Target & EIDO Agent Structuring Task \\
    \midrule
    Incident Type & \texttt{incidentComponent. \newline incidentTypeCommonRegistryText} & Normalize proprietary codes (for example, `211A') to IANA registry value (`ROBBERY-ARMED'). \\
    \midrule
    Initial Problem Description & \texttt{notesComponent. \newline notesActionComments} & Summarize and extract core event narrative from multi-paragraph call-taker notes. \\
    \midrule
    Response Hour, Day, Month & \texttt{callComponent. \newline callStartTimestamp} & Synthesize disparate temporal fields into compliant ISO 8601 timestamp. \\
    \midrule
    Priority Level & \texttt{incidentComponent. \newline incidentCommonPriorityNumber} & Map agency-specific codes (`P1', `Code 3') to standardized numeric scale (1-5). \\
    \midrule
    Sector / Beat & \texttt{locationComponent. \newline locationDescriptionText} & Contextualize jurisdictional areas, appending them to location description. \\
    \midrule
    Call Disposition & \texttt{incidentComponent. \newline incidentDispositionText} & Map disposition codes (`ADV') to descriptive text (`Advised'). \\
    \midrule
    First Unit Arrived & \texttt{resourceStatusComponent. \newline statusTime} & Create distinct status component for specific unit, parsing timestamp. \\
    \midrule
    Unit Time on Scene & \texttt{resourceStatusComponent} fields & Calculate duration; create start and end status objects. \\
    \bottomrule
  \end{tabular}
  \label{tab:transformations}
\end{table}

\subsection{IDX Agent: Incident Correlation and Synthesis}
The IDX Agent is responsible for associating newly created EIDO documents with existing incidents or initiating new incident contexts. This capability is essential in multi-agency and multi-source environments, where a single real-world event generates multiple, partial, and incrementally reported observations.

An incident ($I$) is treated as a set of previously observed EIDO objects:
\begin{equation}
I = \{E_1, E_2, \dots, E_n\}
\end{equation}

When a new EIDO ($E_{new}$) arrives, the agent evaluates its similarity to each existing incident using a weighted scoring function that combines temporal, spatial, and semantic evidence:

\begin{equation}
\Sigma(E_{new}, I) = w_t \cdot \phi_t(\Delta t) + w_g \cdot \phi_g(\Delta g) + w_s \cdot \phi_s(D_E, D_I)
\end{equation}

The three components capture complementary aspects of incident relatedness. The temporal term $\phi_t$ measures how close the new report is in time to recent activity within an incident. The spatial term $\phi_g$ evaluates proximity between reported locations, using straight-line distance by default, with optional alternatives such as network-based or travel-time distance when appropriate data are available. The semantic term $\phi_s$ compares incident descriptions using vector embeddings derived from textual content \cite{reimers2019sentence}.

The weights $w_t$, $w_g$, and $w_s$ control the relative influence of each component and may be learned or configured based on operational context.

The new EIDO is linked to the incident $I^*$ that maximizes the similarity score, calculated using a cascading filter with temporal windows, geospatial proximity, and semantic text similarity, provided that the score exceeds a configurable threshold $\tau$:

\begin{equation}
I^* = \text{argmax}_I \Sigma(E_{new}, I) \quad \text{subject to} \quad \Sigma(E_{new}, I^*) \geq \tau
\end{equation}

The threshold $\tau$ represents the system's tolerance for ambiguity in incident association. Higher values enforce conservative linking, while lower values permit more permissive aggregation. $\tau$ is not specified by the NENA standard and is treated as an operational parameter that may vary by incident type, jurisdiction, or deployment setting. In practice, $\tau$ is configured based on tolerance for false aggregation.

\begin{figure}[ht]
  \centering
  \includegraphics[width=\linewidth]{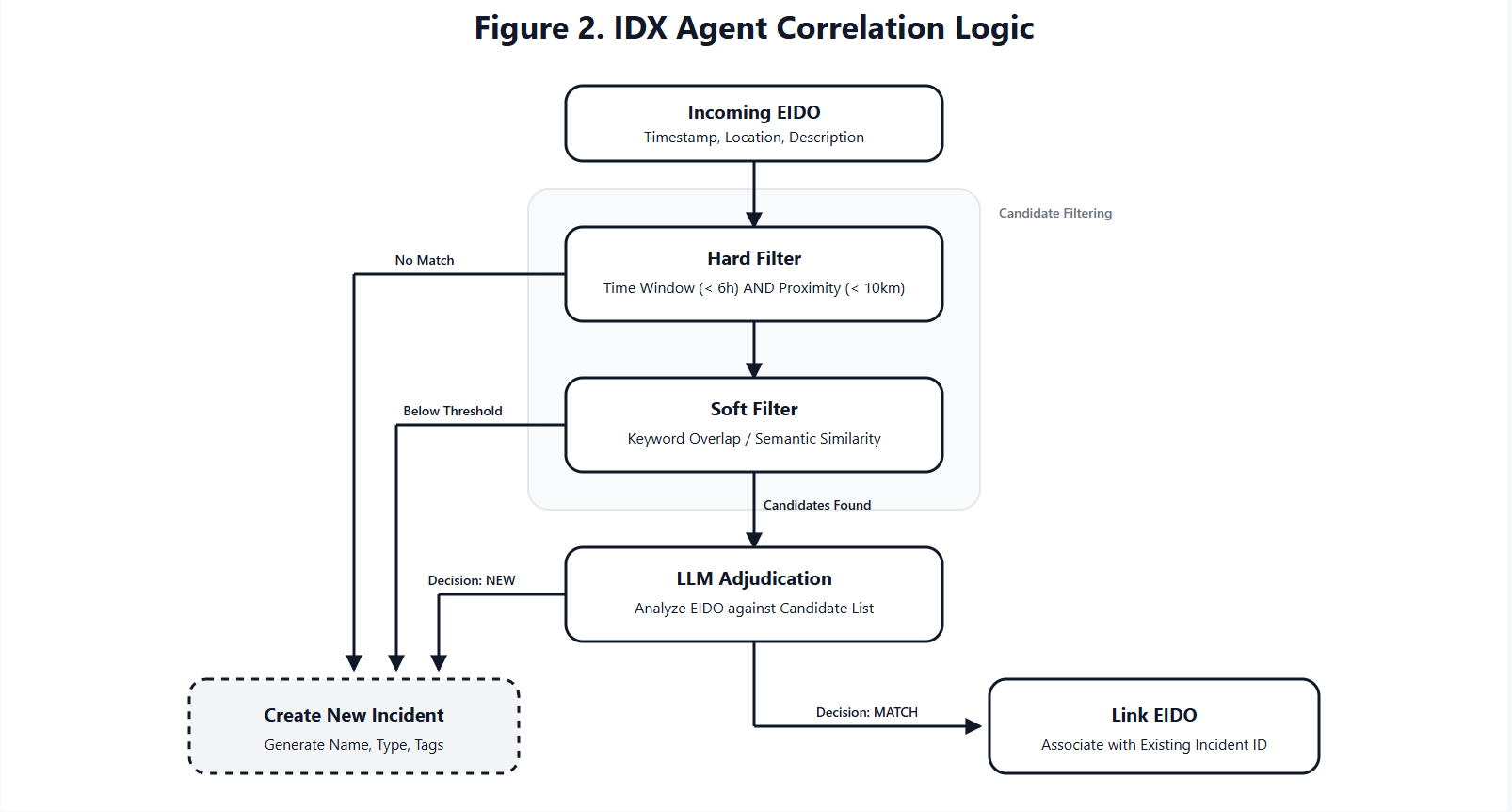}
  \caption{Decision logic employed by the IDX Agent to determine if an incoming report represents a new incident or an update to an existing one.}
  \label{fig:logic}
\end{figure}

Figure \ref{fig:logic} illustrates the decision logic employed by the IDX Agent to determine if an incoming report represents a new incident or an update to an existing one.

If no existing incident satisfies the threshold, the new EIDO initiates a new incident context.

\subsubsection{Composite Representation}
Once individual EIDO objects have been linked to a common incident context, the system can derive a composite representation $C(I)$ that summarizes incident state across all available updates. This composite view is not stored as a separate authoritative record, but is constructed dynamically from the set of linked EIDOs associated with an incident.

Each incident is associated with a collection of linked EIDO objects. Composite attributes are computed by applying aggregation or selection rules over this collection. For example, the set of deployed or involved units $U_C(I)$ is obtained as the union of unit references across all constituent EIDOs:

\begin{equation}
U_C(I) = \bigcup_{i=1}^{n} U(E_i)
\end{equation}

Other incident attributes are derived using rules appropriate to their semantics. Narrative information is constructed by combining \texttt{notesComponent} entries from linked EIDOs, ordered by their timestamps, to provide a consolidated account of reported observations and actions. State-oriented fields, such as \texttt{incidentStatus}, are resolved by selecting the most recent value available among the linked EIDOs, reflecting the current known state of the incident.

This composite representation supports query, visualization, and integration tasks that require a unified view of an incident, while preserving access to the underlying EIDO objects for traceability and audit. By computing composite views from linked updates rather than maintaining a single mutable record, the system avoids loss of information and accommodates the incremental, multi-source nature of emergency incident reporting.

\subsection{Geocoding Agent: Spatial Enrichment}
The Geocoding Agent addresses a common challenge in emergency response: locations are often described using local landmarks, colloquial names, or contextual references rather than formal addresses. Callers may report incidents at a well-known market, a campus landmark, or a locally named site without providing a street address or coordinates.

To resolve such descriptions, the agent combines three complementary processes: leveraging local place knowledge, querying external mapping services, and applying incident context for disambiguation. Unlike general-purpose geocoders and gazetteers, which emphasize formal place names and addresses, the Geocoding Agent is designed to interpret how locations are described in everyday emergency communications. Many locally used names and informal references are absent from standard gazetteers and require additional context to resolve reliably \cite{goodchild2012assuring}.

Resolving a location begins with matching reported place names against a curated collection of locally relevant landmarks, businesses, and commonly used place references. These references reflect local usage and are spatially indexed to support efficient lookup. When local knowledge alone is insufficient, the agent queries public mapping services, such as Google Maps or OpenStreetMap, to retrieve candidate matches that include user-contributed names and informal place descriptions.

When multiple candidate locations are returned, the agent applies available incident context to select the most likely match. Relevant context may include the incident type, the reporting party, the responding jurisdiction, and proximity to other known incident features. By combining these signals, the agent reduces ambiguity without requiring callers to provide fully specified addresses.

The resolved location is incorporated into the EIDO \texttt{locationComponent} in several complementary forms. Geographic coordinates support spatial analysis and mapping, while matched landmark names and descriptive text preserve how the location was originally communicated. Where available, a civic address is included for formal documentation, and a confidence measure records the reliability of the resolution.

\section{FME Integration}
To show how SentinelAI can fit into tools that agencies already use, we connected it to Safe Software's FME \cite{safe_fme}, an enterprise integration platform suitable for spatial data. We use FME as one example of a common data-integration platform. The goal here is not to make FME a requirement, but to show that EIDO-JSON can move in and out of an existing workflow without requiring changes to current operational systems.

FME is widely used in government and public safety to move data between dispatch systems, geographic information systems, databases, and reports. SentinelAI connects to FME through two components: an EIDOReader that turns EIDO-JSON into features inside an FME workspace, and an EIDOWriter that builds valid EIDO-JSON from structured input data. Together, they support two directions of exchange: (i) converting legacy records into EIDO-JSON for SentinelAI to process, and (ii) sending EIDO-JSON to other systems through FME.

\begin{figure}
  \centering
  \includegraphics[width=1\linewidth]{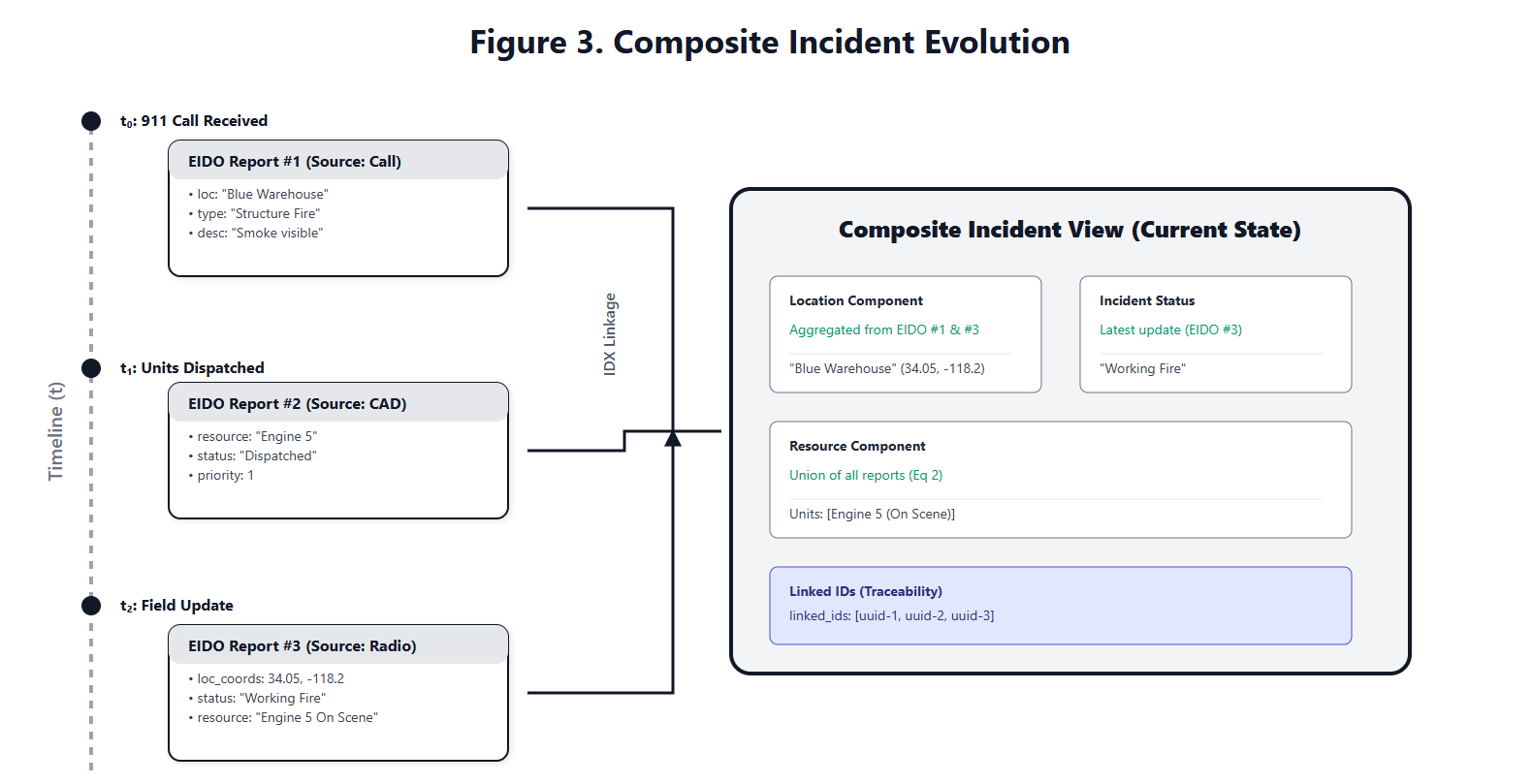}
  \caption{Integration with FME showing EIDOReader and EIDOWriter components facilitating bidirectional data exchange.}
  \label{fig:fme}
\end{figure}

\subsection{EIDOReader}
The EIDOReader loads full EIDO-JSON documents into an FME workspace. It walks through the nested EIDO structure and outputs a set of typed features for major parts of the schema, such as \texttt{incidentComponent}, \texttt{locationComponent}, \texttt{personComponent}, \texttt{resourceComponent}, and \texttt{callComponent}. Field values are copied directly from EIDO-JSON, and links between components are kept using identifiers. This lets FME workflows filter, map, and join incident data using standard table and map operations while keeping the structure of the original EIDO.

\subsection{EIDOWriter}
The EIDOWriter does the reverse: it builds valid, NENA-compliant EIDO-JSON from structured data in an FME workspace (for example, dispatch exports, spreadsheets, or database tables). Input fields are mapped to EIDO components and fields. The writer creates the needed objects, assigns identifiers, and creates the links required by the schema. If the source data is split across several records, the writer can combine them to form a complete EIDO-JSON document. The output can then be shared with NG9-1-1 systems or sent to SentinelAI agents for linking and enrichment.

\section{Illustrative Case Study}
To illustrate the practical application of the SentinelAI framework, we present a small but realistic example where incident information arrives in pieces from separate sources. The key point is that SentinelAI treats each report as its own EIDO-JSON document and then links those documents over time. This matches how incidents evolve in practice and illustrates the linked-update model described in Section 3.2.

\subsection{Sources and Situation}
The example scenario covers heavy rainfall impacts in San Diego County on January 1, 2026. Two sources describe related parts of the same situation.

The first source is a text-based Flash Flood Warning issued by the National Weather Service (NWS) at 12:20 PM PST \cite{nws_flood_2026}. The warning is transmitted as a semi-structured text message:
\begin{quote}
``The National Weather Service in San Diego has issued a Flood Warning for parts of Central San Diego County until tomorrow afternoon, 01/02. Flooding is ongoing or expected to begin shortly due to heavy rain. Do not drive your vehicle through flooded roadways. This storm may intensify, monitor local radio stations and available television stations for additional information and possible warnings from the National Weather Service.''
\end{quote}

The second source is a digital news report published by the San Diego Union-Tribune on the same day \cite{sdut_flood_2026}. This article, titled ``Heavy rainfall floods roadways, knocks out power, causes minor San Diego River flooding,'' provides specific details on infrastructure impacts that complement the official warning.

\subsection{Data Conversion and Structuring}
The SentinelAI pipeline processes these sources independently to generate standardized EIDO documents.

\begin{figure}[ht]
  \centering
  \includegraphics[width=1\linewidth]{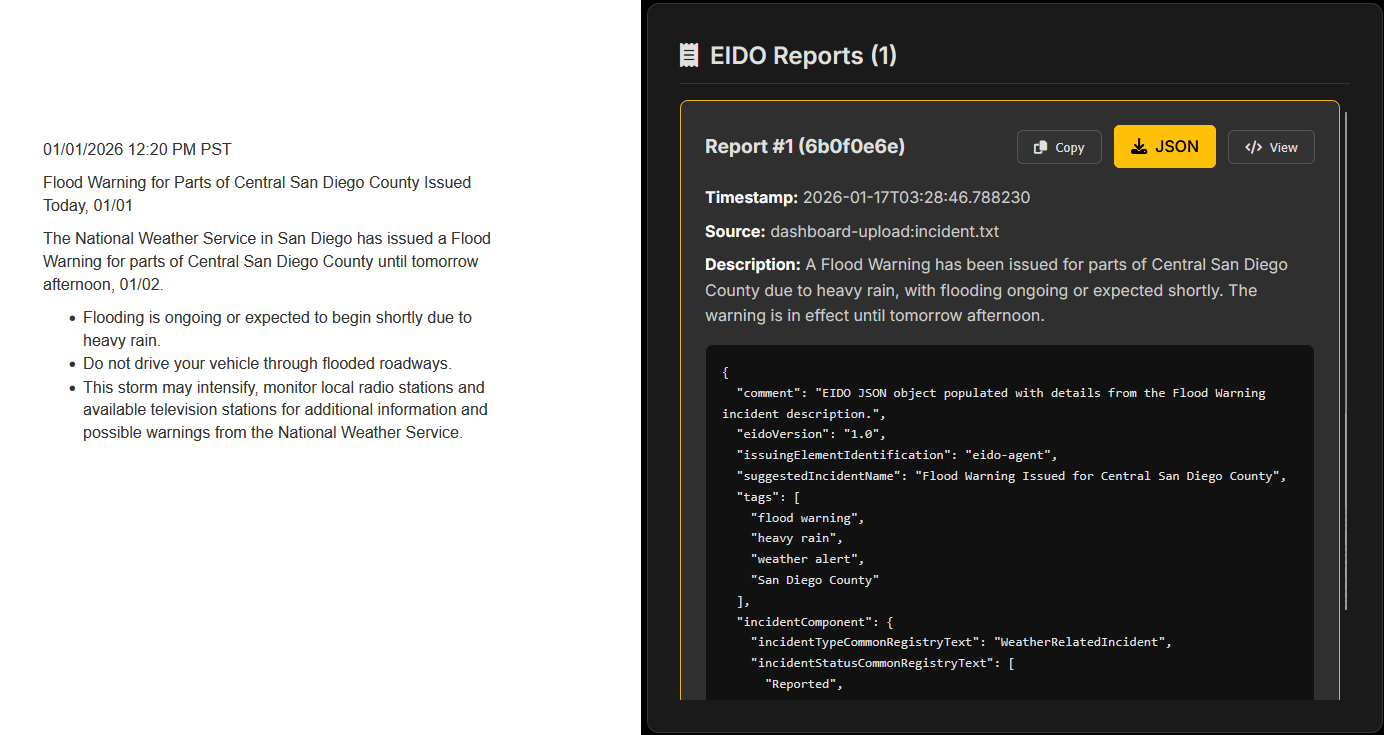}
  \caption{Visual transformation of the NWS Flood Warning text (left) into a structured EIDO-JSON document (right) by the EIDO Agent, highlighting the extracted event type and polygon location data.}
  \label{fig:demo_nws}
\end{figure}

For the NWS warning, the EIDO Agent extracts the event type ``Flood Warning'' and maps it to the registry value \texttt{Weather.Flood}. It identifies the temporal validity window (ending ``tomorrow afternoon'') and resolves the location ``Central San Diego County'' to a polygon representation via the Geocoding Agent. The resulting document, $EIDO_A$, creates an initial incident context (Figure \ref{fig:demo_nws}).

For the news article, the EIDO Agent parses the headline and body text to identify specific impacts. It extracts ``roadway flooding'' and ``power outage'' as \texttt{incidentType} descriptors and identifies the ``San Diego River'' as the primary location. This process produces a second document, $EIDO_B$, which captures the localized impacts reported by the media.

\subsection{Incident Linking}
Once $EIDO_B$ is generated, the IDX Agent evaluates it against the existing incident context created by $EIDO_A$. The agent calculates similarity based on three factors: (1) temporal proximity, as both reports occur on the same day; (2) spatial overlap, as the San Diego River falls within the Central San Diego County polygon; and (3) semantic similarity, as both descriptions share key terms such as ``flood'' and ``rain''.

\begin{figure}[ht]
  \centering
  \includegraphics[width=0.5\linewidth]{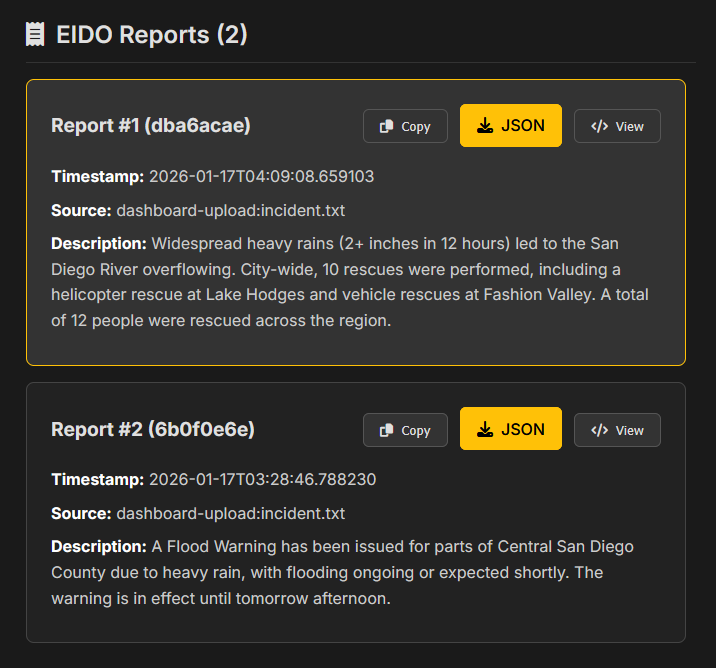}
  \caption{The SentinelAI Dashboard visualization showing the IDX Agent's linking of the NWS Warning and the News Report into a single composite incident timeline.}
  \label{fig:demo_linking}
\end{figure}

The calculated similarity score exceeds the configuration threshold $\tau$, causing the IDX Agent to link $EIDO_B$ to the incident context of $EIDO_A$. The system then presents a composite view of the incident (Figure \ref{fig:demo_linking}) that combines the official warning status from the NWS with the specific impact details provided by the news report, offering a more complete operational picture than either source alone.

This example shows three core behaviors: (i) turning different text sources into NENA-aligned EIDO-JSON (EIDO Agent), (ii) linking separate reports into a shared incident context as updates arrive (IDX Agent), and (iii) forming a composite view from linked updates while keeping the original source documents for review and reuse.

\section{Conclusion}
This paper presented SentinelAI, a multi-agent framework for structuring, linking, and enriching emergency incident data in alignment with the NENA Emergency Incident Data Object standard. The system operationalizes EIDO as an evolving, event-sourced representation, allowing incident information to be updated incrementally as new reports arrive rather than treated as a single static exchange record. Through the coordinated operation of three specialized agents, SentinelAI addresses core challenges in emergency data integration, including interpretation of heterogeneous inputs, correlation of related reports across sources, and spatial enrichment of informally described locations.

The primary contributions of this work are threefold: (1) a reference architecture composed of three specialized agents (EIDO, IDX, and Geocoding); (2) a practical implementation of the NENA EIDO-JSON standard that supports linked incident representations; and (3) a demonstration of enterprise interoperability via FME integration.

Together, these contributions focus on data structuring and interoperability rather than operational decision support. By providing a concrete, standards-aligned implementation and reference architecture, this work aims to support future research and system development that builds on consistent, machine-readable emergency incident data.

\section{Software Availability}
A reference implementation of the SentinelAI architecture, including the EIDO Agent, IDX Agent, Geocoding Agent, and the EIDOReader and EIDOWriter components for FME integration, is available for research and evaluation purposes. Source code, configuration examples, and documentation are provided via a publicly accessible repository: \url{https://github.com/DevKlim/SentinelAI}.

Additional technical details, including representative templates, schema mappings, and illustrative examples, are maintained alongside the reference implementation in the project repository (as accompanying Markdown documentation).

\section*{Acknowledgments}
The EIDO-JSON standard described in this paper is specified by the National Emergency Number Association (NENA) in NENA-STA-024.1.1-2025. We are grateful to Dmitri Bagh from Safe Software for advice on FME integration. Partial support from the US NSF (award 2330460) is gratefully acknowledged.

\bibliographystyle{unsrt}  
\bibliography{references}  

\end{document}